\documentclass{article}
\usepackage{PRIMEarxiv}
\usepackage[utf8]{inputenc} 
\usepackage[T1]{fontenc}    
\usepackage{hyperref}       
\usepackage{url}            
\usepackage{booktabs}       
\usepackage{amsfonts}       
\usepackage{nicefrac}       
\usepackage{microtype}      
\usepackage{lipsum}
\usepackage{fancyhdr}       
\usepackage{graphicx}       
\graphicspath{{media/}}     

\title{Carbon Figures of Merit Knowledge Creation \\with a Hybrid Solution and Carbon Tables API
}

\author{
  Maira de Bayser \\
  IBM Research Brazil \\
  Rio de Janeiro, RJ\\
  \texttt{mgdebayser@br.ibm.com} \\
}

\begin{document}
\maketitle

\begin{abstract}
    Nowadays there are algorithms, methods, and platforms that are being created to accelerate the discovery of materials that are able to absorb or adsorb $CO_2$ molecules that are in the atmosphere or during the combustion in power plants, for instance. In this work an asynchronous REST API is described to accelerate the creation of Carbon figures of merit knowledge, called Carbon Tables, because the knowledge is created from tables in scientific PDF documents and stored in knowledge graphs. The figures of merit knowledge creation solution uses a hybrid approach, in which heuristics and machine learning are part of. As a result, one can search the knowledge with mature and sophisticated cognitive tools, and create more with regards to Carbon figures of merit.
\end{abstract}


\section{Introduction}
Nowadays there are $CO_2$ Capture technologies \cite{sun:2015}\cite{ramdin:2012}, algorithms, methods, and platforms \cite{krallinger:2017} that are being created to accelerate the discovery of materials that are able to absorb or adsorb $CO_2$ molecules that are in the atmosphere. This is needed because of the rapid planet warming\footnote{The United Nations Climate Change Conference (COP26) Portal}, primarily due to burning fossil fuels in power plants that generate greenhouse gas emissions\cite{basso:2021}, for instance. 

According to IUPAC, a \textit{"Polymer is a substance composed of macromolecules, which are molecules of high relative molecular mass, the structure of which comprises the multiple repetition of units derived from molecules of low relative molecular mass"}. And \textit{Permeability}, or \textit{Permeance}, is the most important quantifier of the transport of molecules through polymers. While the \textit{Selectivity} of a Polymer between two types of molecules is the ratio of its permeabilities to those molecules.

In that context, a Figure of Merit (FoM) is a quantifier to the material's effectiveness that is of interest to a material scientist, such as \textit{Permeability}. And to accelerate the creation of Carbon Figures of Merit Knowledge (CFMK), in this work it is presented an asynchronous REST API, called Carbon Tables, that can create the aforementioned knowledge using tables in scientific PDF documents (for instance ,see \cite{kazemian:2017}), and can store it in knowledge graphs. 

There was a need to automatically detect FoM content from tables in published papers, and to create a consolidated table that contains all the figures of merit that are important in the field. For instance, in Membranes Polymers, the most important FoM to us are the $CO_2$ and $N_2$ Permeance in GPU, while also the $CO_2/N_2$ Selectivity.

The main contribution of the Carbon Tables is the consolidation of the results which contain the Carbon FoM knowledge using both a ground truth dataset that I created, and a set of heuristics on the material's name and column headers'.

As a result, the created knowledge graph can be searched with mature and sophisticated AI tools\footnote{IBM Research Deep Search. https://research.ibm.com/interactive/deep-search/}\cite{staar:2018}, and create more knowledge with regards to Carbon FoM. To enable that, another contribution is the asynchronous REST API in which Carbon Tables solution was implemented, which makes the solution high scalable with regards to the number of material scientists' users.

\section{Foundations}

\subsection{Carbon Capture Technologies}

In order to capture Carbon, materials are synthesized using methods, equipment and molecules. A material is a compound that is made of substances. And a membrane, for instance, is a polymer that can adsorb $CO_2$. It has a molar weight, density, viscosity, polarity, melting point, boiling point, solubility, ionization potential, oxidation number, among others properties. 

In the state-of-the-art of Carbon Capture Technologies, there are five worthwhile mentioning: Liquid Absorption, Solid Adsorption, Membranes, Hydrates and Chemical Looping.  

\subsection{Carbon Capture Technologies Figures of Merit}

Each of the Carbon Capture technologies has a set of FoM that may overlap or not between the technologies. For instance, Membranes, Liquid Absorbers and Solid Adsorbers are measured with regards to $CO_2$ sorption capacity. Hydrates effectiveness, on the other hand, is measured with regards to Hydrate mass. While Solid Adsorbers effectiveness is also measured with the adsorption enthalpy, a very important FoM too. 

Publications contain the reactions to synthesize materials that are effective in their corresponding FoM, state variables in which the reactions happen, for instance, temperature, pressure, and pH, and the results of effectiveness.

In Membranes' Technology, for instance, materials' scientists have synthesized a Mixed Matrix Hollow Fiber Membrane (MMHFM) material and measured its performance to adsorb $CO_2$ in a wet lab. The results were published and the FoM are on Table \ref{tab:referencetable}. 

\begin{table}[h!]
\centering
\begin{tabular}{lll}
\hline
Material's Name  & $CO_2$ (GPU) & $CO_2/N_2$ Selectivity \\
\hline
Pure Ultem HFM       & 15.3  & 0.5     \\
MMHFM       & 31.2  & 35.7      \\
\hline
\end{tabular}
\caption{An Exemplar Reference Table, from \cite{kazemian:2017}}
\label{tab:referencetable}
\end{table}

The MMHFM containing functionalized MIL-53 achieved excellent gas permeance and $CO_2/N_2$ selectivity. They also detected that the $CO_2$ Permeance increased for pure Ultem HFM while the ideal $CO_2/N_2$ Selectivity was significantly enhanced simultaneously. According to the authors \cite{kazemian:2017}, $CO_2$ Permeance increased but the $CO_2/N_2$ Selectivity decreased when the temperature increased, which followed the solution-diffusion based transport mechanism. On the other hand, the permeation of MMHFM was similar to that of Pure Ultem HFM, regardless of temperature variations. One can notice then the importance of also detecting the state variables of measurements, such as temperature and pressure.

\section{Carbon Tables Solution Overview}

\begin{figure*}[tb!]
    \centering
    \includegraphics[width=1\textwidth]{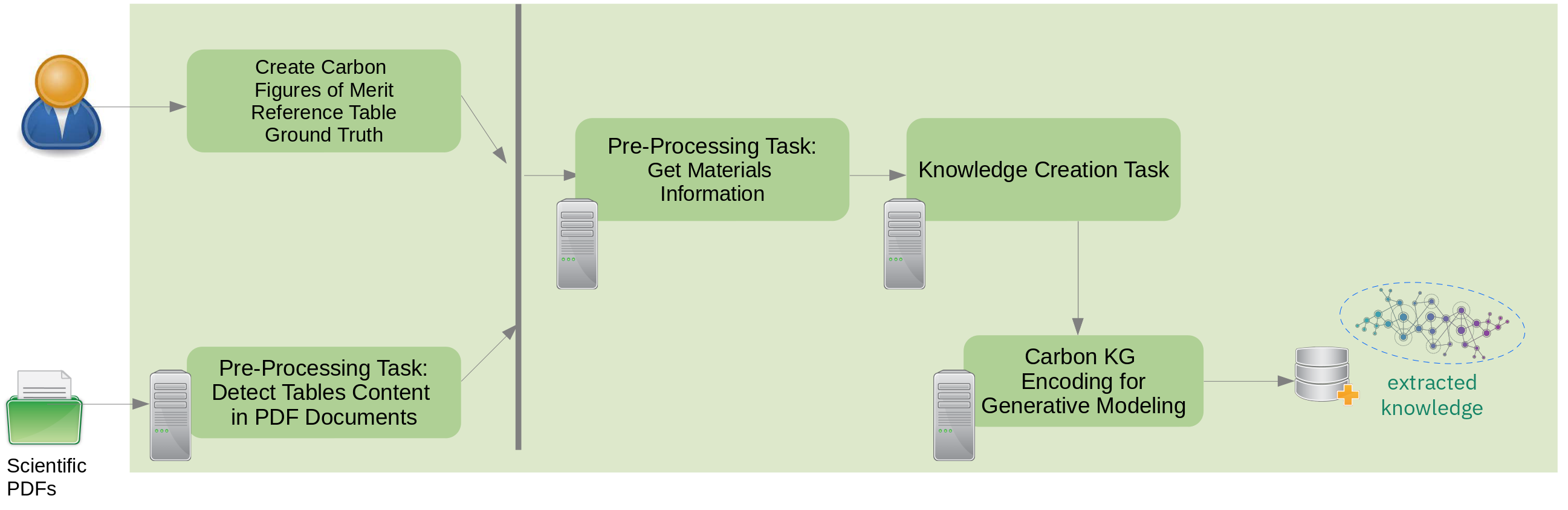}
    \caption{Carbon Tables Solution Overview}
    \label{fig:carbon_tables_solution_overview}
\end{figure*}

The Carbon Tables solution overview is depicted in Figure \ref{fig:carbon_tables_solution_overview}. The data scientist, which can be also the material scientist, creates that herein is called Carbon Figures of Merit Reference Table Ground Truth dataset, or reference table for short. It consists of a document that contains all the materials' name and the figures of merit associated to them with their corresponding values. It may also contain state variables, like temperature and pressure.

Notice that the input of the solution is twofold: the reference table and the annotated tables, which can be done with the pre-processing task: \textit{Detect Tables Content in PDF Documents}.

\begin{itemize}
    \item Pre-processing Task: \textit{Detect Tables Content in PDF Documents}: this task consists of: (i) automatically annotating the papers by: uploading the scientific PDFs into CCS, parsing the PDFs, extracting the tables, downloading the result, and then (ii) finally parsing the annotated results.
    \item Pre-processing Task: \textit{Get Materials Information}: the reference table and annotated tables' results are the input of this task. All materials  that are in the tables are then detected by matching the values of the first column to all materials' names that there are in our Materials knowledge Base (MB).
    \item Knowledge Creation Task: this task consolidates the results which contain the figures of merit knowledge using the reference table, the MB, and a set of heuristics on the material's name, column headers' name, and column values. 
    \item Carbon Knowledge Graph Encoding for Generative Modeling Task: this task consists of encoding the created knowledge into feature vectors that will be the input of a materials design tool using generative modeling.
\end{itemize}

As of today, a possible implementation of the \textit{Detect Tables Content in PDF Documents} is to use the Corpus Conversion Service API\footnote{https://ccs.foc-deepsearch.zurich.ibm.com/docs/} a component of DeepSearch, in the task of detecting the tables in the PDF and extracting their content into JSONs.

The MB can be previously created by adding all the materials information such as SMILES code or Materials synonyms. Information like these could be found in CIRCA\footnote{https://circa.almaden.ibm.com/circa}, the Chemical Information Resources for Cognitive Analytics tool. 

\subsection{Knowledge Creation Task}

\begin{figure*}[ht!]
    \centering
    \includegraphics[width=1\textwidth]{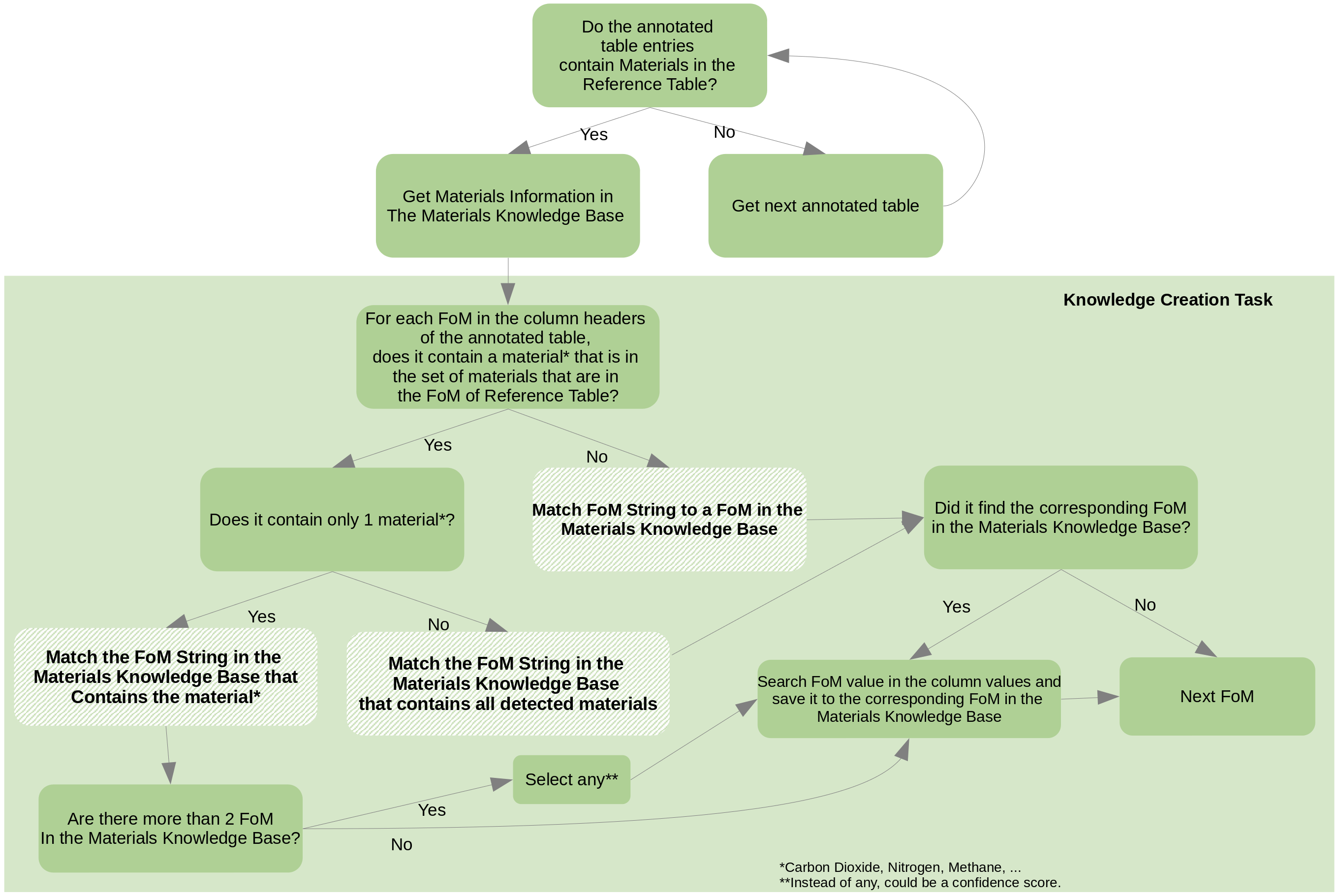}
    \caption{Knowledge Creation Task Decision Tree}
    \label{fig:knowledge_creation_task}
\end{figure*}

Once the previous tasks are complete, the Figure of Merit Knowledge Creation task starts. It uses as the input the reference table, the MB, and Figure \ref{fig:knowledge_creation_task} illustrates how the knowledge is created.

The decision tree starts as the following: for each of the annotated table, if all entries contain Materials that are in the reference table, get the next table. Otherwise get all materials information in the MB. 

Then for each FoM in the column headers of the annotated table, verify if the column header contains a material, for instance $CO_2$ or $N_2$, that is in the set of materials that are in the FoM of reference table. If it contains and it contains only one material, the system matches all the FoM String in the MB that contains that material. Otherwise, the system matches the FoM String, and the system searches the FoM value in the column values. Once found, the system saves it to the corresponding FoM in the MB, and do it again to the next FoM in the table.

If more than 2 FoMs are matched, for instance Permeability of $CO_2$ and Selectivity of $CO_2$/$N_2$, the system selects the first, and again the system searches the FoM value in the column values.

If the system did not find a material in the column header, the system matches the FoM String to a FoM in the MB. If it finds the corresponding FoM in the MB, it then searches the FoM value in the column values and saves it. Once that is done to all FoMs of all column headers, the system gets the next annotated table.

In this decision tree, the following steps are highlighted:
\begin{itemize}
    \item Match FoM String to a FoM in the Materials Knowledge Base:\\
    For instance, if the figure of merit does not contain any material, such as $CO_2$ in the String, then the match is a simple search of Strings with no ambiguity. 
    \item Match the FoM String in the Materials Knowledge Base that contains the material:
    If the figure of merit does contain a material, such as $CO_2$, the system matches all Strings using regular expression. However, an Support Vector Machine-based system (or any supervised machine learning-based algorithm) could be used in which the Strings are encoded into vectors and the accuracy of the match is increased and to solve the ambiguity.
    \item Match the FoM String in the Materials Knowledge Base that contains all detected materials:
    If the figure of merit does contain more than one material, such as $CO_2$ and $N_2$, the system matches all Strings using regular expression. In our reference table all figures of merit with more than one material are unique instances, i.e, there aren't different figures of merit with the same two or more materials. Therefore it would be always one match.
\end{itemize}

The MB also contains dictionaries to the materials, FoMs, etc. For instance, the absorption capacity can also be mentioned in the tables as absorption flux.

\section{Carbon Tables API}

The Carbon Tables API implemented is an Asynchronous Client/Server API that runs on top of AIOHTTP \cite{aiohttp}\cite{wilkes:2020} and Ray \cite{ray}\cite{ray:2022}. This was needed to make the solution scalable to run on IBM Cloud with containers, such as Kubernetes, and on OpenShift cluster.

In our Carbon Tables solution, the Carbon Reference Table Ground Truth dataset herein presented contains, for instance, 238 entries values, temperature and pressure information (in which the adsorption or capacities were measured), and the FoMs are mainly about Membranes, Liquid Absorption, Solid Adsorption, and Hydrates. To create this dataset, all the FoMs were manually detected on the tables, for instance on the following subset of papers: \cite{kazemian:2017} \cite{xu:2019} \cite{huertas:2017} \cite{shi:2019} \cite{ramezani:2018}  \cite{babaei:2019} \cite{thirion:2016} \cite{yang:2012} \cite{legoix:2018}.

It contains the SMILES code of the material, which is a String, it contains four common fields to all Carbon Capture technologies, 19 fields to Liquid/Solid sorption, 3 fields to Hydrates, 14 fields to Membranes and 9 fields to Chemical Looping. All of these fields are floats. There is also the category field which is a list.

The dataset was then converted into the Materials Knowledge Base (MB) as a Class in Python that is used in the API and contains all materials and their information. 

When the user calls the Carbon Tables API, the user can: \\
(i) select a file in disk that can be a PDF or a Zip file that contains all PDFs that contain the tables; \\(ii) upload and parse the file(s) selected in step (i); \\(iii) call the API to create the Carbon figures of merit knowledge with the consolidation task, \\(iv) download the created knowledge in step (iii) once the task is done; \\
(v) search the results using libraries like, for instance, Pandas \cite{pandas}\cite{pandas:2022}.

\begin{figure*}[h]
    \centering
    \includegraphics[width=1\textwidth]{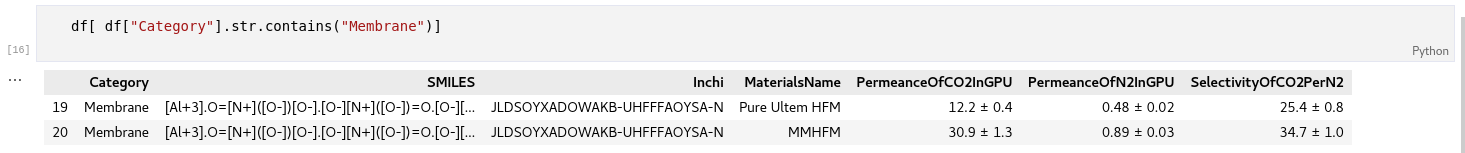}
    \caption{Query of Figures of Merit in the Membrane Technology Category with Pandas.}
    \label{fig:query}
\end{figure*}

As an example on how to query the results using Pandas, Figure \ref{fig:query} contains the Python code in which the user can search the created knowledge for FoM in the \textit{Membrane} technology \textit{Category}. These FoMs were in Table \ref{tab:referencetable} from paper \cite{kazemian:2017}, described in Section 2.

\section{Related Work}

Algorithms have been applied to make use of content derived directly from documents to recognize Chemical Entities, or in other words, to model how chemical and drug names mentions are semantically different from the surrounding text. The state-of-the art though \textbf{\textit{does not contain solutions to learn materials' figures of merit from tables}} in documents nor using Supervised or Unsupervised Machine Learning methods. Possibly because there is a lack of corpus for Carbon Capture.

Among the Supervised Machine Learning methods for Chemical Knowledge Recognition, there are Random Forests that were applied to NER tasks  \cite{lammurias:2015} in a corpus of MEDLINE titles and abstracts, a biomedical database resource, that were manually annotated by domain experts. To each recognized chemical entity, they associated a chemical entity in a ontology and calculated a validation score based on its similarity of other terms using a hybrid approach: Conditional Random Fields and Random Forests. They took into account only the most relevant ancestors of a concept in the ontology, and they used the h-index to measure the relevance.

In \cite{jessop:2011}\cite{corbett:2008}, five classes of named entity (compound, reaction, adjective, enzyme and prefix) of chemistry papers were annotated, and the solution was tested on 500 annotated PubMed abstracts and titles.
Three classifiers were used to recognise chemical names, in which Maximum-Entropy Markov Models were used, with a handcrafted feature set, to estimate the conditional probability of tag sequences corresponding to named entities.

In \cite{ponomareva:2007}, Biomedical Named-Entity Recognizers were built with Hidden Markov Models and Conditional Random Fields (CRD), and were evaluated using a MEDLINE corpus (titles and abstracts) with 5 classes of biomedical entities: protein, RNA, DNA, cell type and cell line.

After encoding the tokenized terms using a set of taggers, including POS-tagger, and a Bio tagger, Support Vector Machines (SVM) and CRF were applied as classifiers and evaluated as NERs \cite{tang:2015} in a abstract corpus of chemistry-related disciplines with eight types of chemical entities.

Mixed CRFs have been used as a method to solve a sequence labeling problem in chemical compound and drug name recognition \cite{lu:2015} in PubMed corpus. The authors hand-crafted a tokenizer, encoded the words with Mikolov's Word Embedding, clusterized the words with K-means, and then used the multi-level as word features in the mixed CRFs to tag the chemical names.

In \cite{gajendran:2020} Biomedical Named Entity Recognition (BNER) is the task of extracting chemical names from biomedical texts to support biomedical and translational research. The BioCreAtIvE II GM corpus is used and the top attained model is used to run on JNLPBA and NCBI corpus. Four different variants were evaluated: LSTM with word embedding, Bidirectional LSTM with word embedding, Bidirectional LSTM with character embedding, and finally the Bidirectional LSTM with character and word level embedding as well as CRF. Experimental results showed that the BLSTM-RNN model with the embeddings achieved better performance compared to the existing state-of-the-art systems.

\section{Conclusions and Future Work}

In this work I presented an asynchronous REST API to accelerate the creation of Carbon figures of merit knowledge using tables in scientific PDF documents. The knowledge is stored in knowledge graphs. Our solution uses a hybrid approach, in which heuristics and machine learning are part of. As a result, one can search the knowledge with mature and sophisticated cognitive tools, and create more with regards to Carbon figures of merit.

The main contribution of the Carbon Tables is the consolidation of the results which contain the Carbon FoM knowledge using both a ground truth dataset that we created, and a set of heuristics on the material's name and column headers'. Another contribution is the asynchronous REST API in which Carbon Tables solution was implemented, which makes the solution high scalable with regards to the number of material scientists' users.

As an on going work, the hybrid solutioncan be enhanced with the usage of a bigger corpus in which learned models will do the majority of the FoM knowledge creation task.

\section*{Acknowledgements}
I greatefully acknowledge support from Maximilien de Bayser, and Ronaldo Giro from IBM Research Brazil, and Christoph Auer from IBM Research Zurich, during the implementation of the solution presented in this paper.

\bibliographystyle{unsrt}  
\bibliography{co2_arxiv}  

\end{document}